\documentclass[11pt]{article}

\usepackage[final]{acl}
\usepackage{times}
\usepackage{latexsym}

\usepackage[T1]{fontenc}
\usepackage[utf8]{inputenc}

\usepackage{microtype}

\usepackage{inconsolata}
\usepackage{makecell} 

\usepackage{graphicx}
\usepackage{amsmath}
\usepackage{booktabs}
\usepackage{amssymb}

\title{PrunePath: Towards Highly Structured Sparse Language Models}

\author{Zhexuan GU \and Zixun FU \and Yancheng Yuan\thanks{Corresponding author.} \\
        Department of Applied Mathematics, The Hong Kong Polytechnic University\\  
        \texttt{\{zhexuan.gu, zixun.fu\}@connect.polyu.hk},
        \texttt{yancheng.yuan@polyu.edu.hk}}

\begin{document}
\maketitle
\begin{abstract}
Feed-forward networks (FFNs) dominate the parameter count and computation of modern language models, yet existing pruning methods often struggle to convert sparsity into hardware-friendly inference efficiency gains. We introduce \textbf{PrunePath}, a budget-adaptive structured sparsification framework for FFN layers. Built on MoEfication, PrunePath replaces independent expert-wise thresholding with a softmax-normalized routing distribution and activates important experts under a cumulative-mass threshold. This formulation imposes a token-level probability budget, enabling adaptive expert counts and a direct inference-time sparsity knob from a single checkpoint. Across NLU, NLG, and instruction-tuning evaluations, PrunePath achieves a favorable sparsity--performance trade-off compared with existing static pruning and MoEfication-based methods.
 We further implement Triton kernels for KV-cache decoding to translate the resulting structured sparsity into practical memory savings and measurable decoding-speed improvements. These results demonstrate the superior performance of PrunePath for building highly sparse, deployment-friendly large language models.
\end{abstract}

\section{Introduction}
\label{sec:introduction}
The exponential scaling of large language model (LLM) parameters has driven remarkable advances across diverse tasks, from natural language understanding (NLU) to complex reasoning in mathematics and code generation~\cite{brown2020language,yang2024qwen2technicalreport}. However, this massive parameterization imposes a severe memory wall: the static footprint for model residency and the transient peak memory from intermediate activations together constrain deployment, especially on resource-limited devices~\cite{li2025tpi}.

Significant strides have been made in optimizing the attention module~\cite{vaswani2017attention}. From a systems perspective, the FlashAttention series~\cite{dao2022flashattention,dao2024flashattention} eliminates the materialization of large intermediate tensors through IO-aware execution. Algorithmically, sparse attention~\cite{beltagy2020longformer,xu2025xattention} and linear attention~\cite{katharopoulos2020transformers} decouple the quadratic complexity from sequence length. At the deployment level, KV-cache management techniques such as H2O~\cite{zhang2023h2o} and efficient serving engines~\cite{kwon2023efficient,zheng2024sglang} further reduce memory overhead.

\begin{figure}
    \centering
    \includegraphics[width=0.85\linewidth]{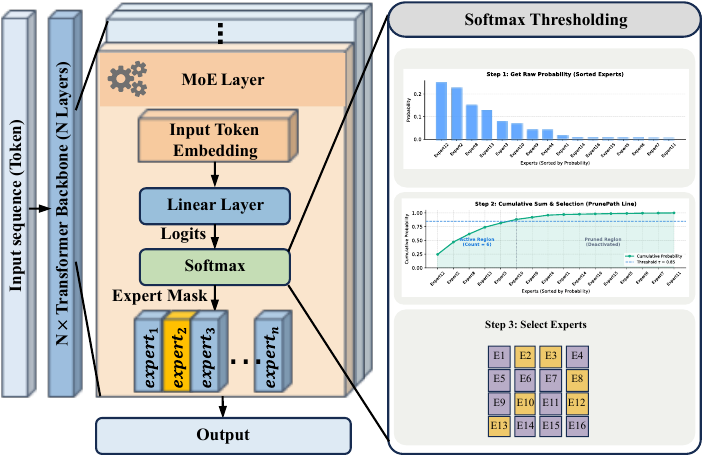}
    \caption{PrunePath visualization.}
    \label{fig:prune_path}
\end{figure}

Compared to the systematic optimization of attention, the pursuit of an ideal balance between FFN efficiency and hardware-friendly deployment remains an open challenge. The FFN typically accounts for two-thirds of total parameters~\cite{wang2022finding} and dominates computational FLOPs under typical sequence lengths~\cite{kaplan2020scaling}. The large matrix multiplications within FFN layers materialize high-dimensional intermediate activations, creating a peak memory bottleneck that is particularly acute on edge devices.

To address the FFN bottleneck, various pruning strategies have been explored. Early efforts primarily focused on fine-grained pruning, employing either heuristic-based neuron importance~\cite{han2016deep, sun2024simple} or optimization-based methods leveraging second-order derivative information~\cite{lecun1989optimal, frantar2023sparsegpt} to sparsify weight matrices.
However, the resulting unstructured sparsity fails to translate into tangible inference speedups due to irregular memory access. The hardware-friendly N:M sparsity pattern~\cite{mishra2021accelerating} addresses this irregularity, yet its rigid structure often incurs non-negligible performance degradation compared to fine-grained counterparts.

A more promising direction toward structured compression is MoEfication~\cite{zhang2022moefication}, which partitions dense FFN layers into a Mixture-of-Experts (MoE) structure and activates only a subset of experts per token during inference. Although the total model size is preserved, the reduced expert dimensionality lowers both peak memory and per-token FLOPs. The sparse activation principle underlying MoE has proven highly effective in 
production-grade models such as Mixtral~\cite{jiang2024mixtral} and DeepSeek-V3~\cite{liu2024deepseek}.

\begin{figure}[t]
    \centering
    \includegraphics[width=0.49\textwidth]{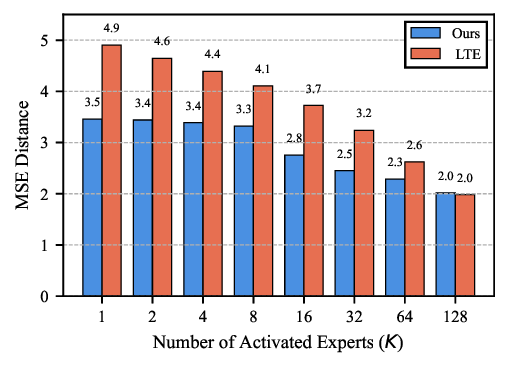}
    \caption{
    \textbf{Motivating top-$k$ reconstruction analysis.}
    We compare LTE and PrunePath by retaining only the top-$k$ ranked experts and measuring the MSE to each method's own all-expert reference output. }
    \label{fig:mse_comparison}
\end{figure}

Building on MoEfication, Learn-to-be-Efficient (LTE)~\cite{zheng2024learn} represents the current state-of-the-art method. LTE clusters FFN neurons into expert groups and trains sigmoid-based routers that independently score each expert, activating those whose scores exceed a predefined threshold. While this design avoids using softmax-normalized weights for output aggregation, it treats expert activation as a set of independent binary decisions. As a result, the number of activated experts is controlled only indirectly by the threshold, which may lead to conservative over-activation. As illustrated in Figure~\ref{fig:mse_comparison}, LTE exhibits larger reconstruction error when only a small number of top-ranked experts are retained.

To bridge this gap, we present \textbf{PrunePath} as visualized in Figure~\ref{fig:prune_path}, a budget-adaptive sparse FFN framework that activates a set of important experts under a cumulative-mass threshold $\tau$. This routing strategy introduces competition among experts under a token-level global probability budget, providing a direct knob for controlling token-wise sparsity. Extensive evaluations across diverse benchmarks demonstrate that PrunePath achieves substantial sparsity while maintaining competitive task performance over practical sparsity ranges.
Our contributions are as follows:

\begin{itemize}
    \item \textbf{Cumulative-Mass Competitive Routing.}
    PrunePath replaces independent sigmoid-threshold routing with a softmax-normalized expert distribution and activates top-ranked experts controlled by a cumulative probability threshold. By imposing a token-level global probability budget, PrunePath enables adaptive per-token expert counts and mitigates the potential conservative over-activation.

    \item \textbf{Single-Checkpoint Dynamic Sparsity.} 
    PrunePath's progressive training yields a single checkpoint with an inference-time sparsity knob. By adjusting the cumulative-mass threshold $\tau$, the same checkpoint traces a smooth efficiency--accuracy frontier over a practical range of sparsity levels, enabling flexible efficiency--accuracy trade-offs without retraining or checkpoint switching.

    \item \textbf{Triton-Accelerated Sparse FFN Inference.} 
    We implement custom Triton~\cite{tillet2019triton} kernels for KV-cache autoregressive decoding that translate structured FFN sparsity into substantial peak-memory reductions and decode-only latency improvements.
\end{itemize}

\section{Preliminaries}
\label{sec:preliminary}

\subsection{Feed-Forward Networks in Modern LLMs}

Modern LLMs predominantly adopt gated FFN architectures 
with bias-free linear projections~\cite{chowdhery2023palm, yang2024qwen2technicalreport}. 
A representative formulation is the SwiGLU variant~\cite{shazeer2020glu}:
\begin{equation}
    \label{eq:ffn}
    \mathrm{FFN}(x) = \left(\phi(xW_{\mathrm{gate}}) \odot xW_{\mathrm{up}}\right) W_{\mathrm{down}},
\end{equation}
where $x\in \mathbb{R}^{d}$ is a hidden representation for a token, $W_{\mathrm{gate}}, W_{\mathrm{up}} \in \mathbb{R}^{d \times d_{\mathrm{ff}}}$, 
$W_{\mathrm{down}} \in \mathbb{R}^{d_{\mathrm{ff}} \times d}$ are weight matrices, $\phi(\cdot)$ 
denotes the activation function (e.g., SiLU), and $\odot$ is the Hadamard product. 
The key structural property is that the intermediate representation 
$h = \phi(xW_{\mathrm{gate}}) \odot xW_{\mathrm{up}} \in \mathbb{R}^{d_{\mathrm{ff}}}$ 
is computed element-wise along the intermediate dimension $d_{\mathrm{ff}}$, 
making intermediate neurons separable along this dimension. The analysis below 
extends naturally to any FFN variant that preserves this separability.

\subsection{MoEfication}

MoEfication~\cite{zhang2022moefication} decomposes a dense FFN into an 
equivalent MoE by clustering neurons along the intermediate dimension. Specifically, balanced $k$-means clustering~\cite{bradley2000constrained} is applied to intermediate-neuron weights to obtain a permutation matrix
$\Pi \in \{0,1\}^{d_{\mathrm{ff}} \times d_{\mathrm{ff}}}$ 
that groups similar neurons into contiguous blocks. Defining the permuted weights as
\begin{equation*}
\small
    \label{eq:permuted_weights}
    \begin{aligned}
        \tilde{W}_{\mathrm{gate}} = W_{\mathrm{gate}}\,\Pi, 
        \tilde{W}_{\mathrm{up}}   = W_{\mathrm{up}}\,\Pi, 
        \tilde{W}_{\mathrm{down}} = \Pi^{\top}W_{\mathrm{down}},
    \end{aligned}
\end{equation*}
the intermediate dimension $d_{\mathrm{ff}}$ is evenly partitioned into $E$ 
experts, each of width $d_e = d_{\mathrm{ff}} / E$. Since the Hadamard product 
operates element-wise along the intermediate dimension, we know that
\begin{equation}
    \label{eq:moefication_decomp}
    \begin{array}{ll}
    \mathrm{FFN}(x) 
    & = \left(\phi(xW_{\mathrm{gate}}) \odot xW_{\mathrm{up}}\right) \Pi\Pi^{\top} W_{\mathrm{down}} \\
    & = \sum\limits_{e=1}^{E} \underbrace{
        \left(\phi\!\left(x\tilde{W}_{\mathrm{gate}}^{(e)}\right) 
        \odot\, x\tilde{W}_{\mathrm{up}}^{(e)}\right) 
        \tilde{W}_{\mathrm{down}}^{(e)}
    }_{\displaystyle \mathrm{FFN}_{e}(x)},
    \end{array}
\end{equation}
where $\tilde{W}_{\mathrm{gate}}^{(e)}, \tilde{W}_{\mathrm{up}}^{(e)} 
\in \mathbb{R}^{d \times d_e}$ and $\tilde{W}_{\mathrm{down}}^{(e)} 
\in \mathbb{R}^{d_e \times d}$ are the sub-matrices of the $e$-th expert. 
Crucially, Eq.~\eqref{eq:moefication_decomp} is an \emph{identity 
transformation}: activating all $E$ experts recovers the original dense 
FFN output exactly. The goal of subsequent expert routing is therefore to identify a small activated set $\mathcal{A} \subset \{1,\dots,E\}$ such that the aggregation of selected expert outputs approximates the dense FFN output while reducing computation.

\subsection{Expert Routing in LTE}

LTE~\cite{zheng2024learn} trains a lightweight router to determine which experts to activate. 
For each expert $e$, the router computes an independent score through a sigmoid gate:
\begin{equation}
    \label{eq:lte_gate}
    G_e(x) = \mathrm{sigmoid}\!\left(w_e^{\top} x\right),
\end{equation}
where $w_e \in \mathbb{R}^d$ is a learnable routing vector.

LTE employs a two-stage training procedure. In the first stage, it uses a soft routing mode with all experts activated and weighted by sigmoid scores:
\begin{equation}
    \label{eq:lte_soft}
    y_{\mathrm{soft}} = \sum_{e=1}^{E} G_e(x) \cdot \mathrm{FFN}_e(x).
\end{equation}
In the hard routing mode used for model adaptation and inference, experts are selected by an expert-wise threshold:
\begin{equation}
    \label{eq:lte_hard_set}
    \mathcal{A}_{\mathrm{LTE}}(x) = \{e \mid G_e(x) > \delta\},
\end{equation}
and the layer output is computed by aggregating the selected expert outputs:
\begin{equation}
    \label{eq:lte_hard_output}
    y_{\mathrm{hard}} = \sum_{e \in \mathcal{A}_{\mathrm{LTE}}(x)} \mathrm{FFN}_e(x),
\end{equation}
where $\delta$ is a predefined threshold.

Although LTE uses an efficiency penalty to encourage sparse routing during training, inference-time activation is still determined by independently thresholding each expert score. Thus, the active expert count is controlled only indirectly by the learned score distribution and the threshold $\delta$.

\section{Our Method}
\label{sec:method}

We present \textbf{PrunePath}, a budget-adaptive sparse FFN framework built on the MoEficated FFN decomposition. 
Instead of making independent binary decisions for each expert, PrunePath constructs a normalized routing distribution and activates important experts under a prescribed cumulative-mass budget. 
This cumulative-mass routing provides a direct sparsity knob through the threshold $\tau$ and enables adaptive per-token expert counts.

\subsection{Cumulative-Mass Expert Activation}
\label{sec:method_exp_activation}

Given a token representation $x \in \mathbb{R}^{d}$, a lightweight linear router with weight $W_r \in \mathbb{R}^{d \times E}$ produces expert logits
\begin{equation}
    \label{eq:routing_logits}
    g = W_r^\top x \in \mathbb{R}^{E},
\end{equation}
where $E$ is the number of experts. We normalize the logits with softmax:
\begin{equation}
    \label{eq:softmax_prob}
    p_j = \frac{\exp(g_j)}{\sum_{i=1}^{E}\exp(g_i)}, 
    \quad j=1,\dots,E.
\end{equation}

Let $\pi$ denote the descending order of expert probabilities, i.e.,
$p_{\pi_1} \geq p_{\pi_2} \geq \cdots \geq p_{\pi_E}$. 
PrunePath activates the high-probability experts whose cumulative mass remains below a threshold $\tau$, while always retaining the top-ranked expert:
\begin{equation}
    \label{eq:tau_thresh_activate}
    \mathcal{A}_{\tau}(x)
    =
    \{\pi_1\}
    \cup
    \left\{
    \pi_i \;\middle|\;
    \sum_{r=1}^{i}p_{\pi_{r}} < \tau
    \right\}.
\end{equation}
This selection rule is inspired by top-$p$ nucleus sampling~\cite{Holtzman2020The}, in the sense that both operate on a probability-sorted sequence controlled by cumulative probability mass.

For output aggregation, we adopt sigmoid-gated expert weighting, analogous to the soft
routing stage of LTE~\cite{zheng2024learn}, but apply it only to the experts selected by the cumulative-mass rule:
\begin{equation}
    \label{eq:moe_output}
    y_{\tau}(x)
    =
    \sum_{e \in \mathcal{A}_{\tau}(x)}
    \mathrm{sigmoid}(g_e) \cdot \mathrm{FFN}_{e}(x).
\end{equation}
This decouples competitive expert selection from output scaling: softmax imposes a global probability budget for deciding which experts to execute, while sigmoid gates provide independent, non-normalized scaling for selected expert outputs.

\subsection{Progressive Sparsity-Path Training}
\label{sec:method_train}

Directly training under an aggressive sparsity target can be unstable, since many experts receive limited task signal once excluded by the hard routing mask. 
PrunePath first uses an all-expert warm-up stage with
$\tau_{\mathrm{warm}}=1.05$, so that all experts are activated. It is followed by a progressive sparsity path and
decrease the cumulative-mass threshold toward a target value $\tau_{\min}$:
\begin{equation}
    \label{eq:tau_schedule}
    \tau_t
    =
    1.0 - (1.0-\tau_{\min})\frac{t}{T},
    \quad t=1,\ldots,T,
\end{equation}
where $T$ is the number of training rounds. 
As $\tau_t$ decreases, the model is progressively exposed to sparser expert subsets. 

To make cumulative-mass pruning effective, we encourage token-level routing distributions to be sharp. 
Given a batch of $B$ token representations, let $p_{ij}$ be the softmax routing probability from token $i$ to expert $j$. We minimize the entropy loss:
\begin{equation}
    \label{eq:entropy_loss}
    \mathcal{L}_{\mathrm{ent}}
    =
    -\frac{1}{B}
    \sum_{i=1}^{B}
    \sum_{j=1}^{E}
    p_{ij}\log(p_{ij}+\epsilon),
\end{equation}
where $\epsilon$ is a small constant for numerical stability.

Entropy minimization alone may collapse routing mass onto a few experts. 
We therefore add a batch-level load balancing loss:
\begin{equation}
    \label{eq:load_balance_loss}
    \mathcal{L}_{\mathrm{bal}}
    =
    E \sum_{j=1}^{E}
    \left(
    \frac{1}{B}\sum_{i=1}^{B}p_{ij}
    \right)^2.
\end{equation}
The entropy loss encourages token-level specialization, while the load balancing loss prevents global expert collapse.

The final objective is:
\begin{equation}
    \label{eq:train_loss}
    \mathcal{L}
    =
    \mathcal{L}_{\mathrm{task}}
    +
    \eta \mathcal{L}_{\mathrm{ent}}
    +
    \lambda \mathcal{L}_{\mathrm{bal}},
\end{equation}
where $\eta$ and $\lambda$ control the strengths of entropy minimization and load balancing.

\section{Experiment}
\label{sec:experiment}

\subsection{Experimental Settings}

\textbf{Models and Datasets.}
We evaluate PrunePath across NLU, NLG, and instruction-tuning settings. The comprehensive configurations of models and benchmarks are summarized in Table~\ref{tab:models_datasets}.

\begin{table}[htbp]
\centering
\caption{Overview of evaluation settings, models, and datasets.}
\label{tab:models_datasets}
\vskip 0.1in
\begin{small}
\resizebox{0.5\textwidth}{!}{
\begin{tabular}{lll}
\toprule
\textbf{Setting} & \textbf{Models} & \textbf{Datasets} \\
\midrule
NLU & \makecell[l]{RoBERTa-large \\ \cite{liu2019roberta}} & \makecell[l]{SST-2 \cite{sst2} \\ MNLI \cite{mnli} \\ QNLI \cite{qnli} \\ MRPC \cite{mrpc}} \\
\addlinespace
NLG & \makecell[l]{GPT-2 Medium \\ \cite{gpt2} \\  Pangu-1B \\ \cite{pangu}} & \makecell[l]{XSum \cite{xsum} \\ WikiText \cite{wikitext}} \\
\addlinespace
SFT & \makecell[l]{Qwen2-7B \\ \cite{yang2024qwen2technicalreport}} & \makecell[l]{Fine-tuning: tulu-v2 \cite{tulu} \\ Eval: MMLU \cite{mmlu1,mmlu2}} \\
\bottomrule
\end{tabular}
}
\end{small}
\vskip -0.1in
\end{table}

We report FFN sparsity as the primary efficiency metric following LTE. For an input containing
\(N\) tokens and a model with \(L\) FFN layers, let \(\mathcal{A}_{\ell}(x_i)\)
denote the set of experts activated for token \(x_i\) at layer \(\ell\). We
define the average FFN neuron sparsity as
\begin{equation}
    s_{\mathrm{FFN}}
    =
    1 -
    \frac{1}{NLE}
    \sum_{i=1}^{N}
    \sum_{\ell=1}^{L}
    |\mathcal{A}_{\ell}(x_i)|.
\end{equation}
For dynamic MoEfication-based methods, this metric measures the average fraction
of inactive FFN experts. For Wanda, we report the target FFN weight sparsity of
its static pruning mask.

\textbf{Baselines.}
We compare PrunePath with two representative pruning baselines: 
(a) LTE~\cite{zheng2024learn}, a strong MoEfication-based pruning method which serves as our main baseline; (b) Wanda~\cite{sun2024simple}, a widely used post-training weight pruning method.

\subsection{Results and Analysis}

\subsubsection{Performance on NLU Tasks}

\begin{figure}[t]
    \centering
    \includegraphics[width=0.49\columnwidth]{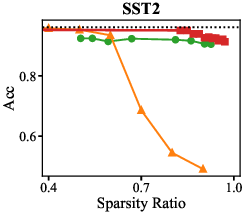}
    \includegraphics[width=0.49\columnwidth]{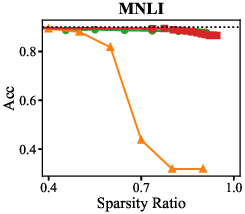}
    \includegraphics[width=0.49\columnwidth]{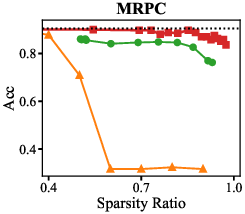}
    \includegraphics[width=0.49\columnwidth]{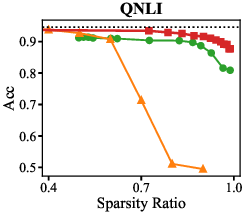}
    \includegraphics[width=1.0\columnwidth]{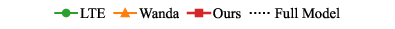}
    \caption{NLU results with RoBERTa-large. PrunePath maintains stronger accuracy over a
        wider FFN sparsity range than Wanda and LTE.}
    \label{fig:nlu_results}
\end{figure}

We show the performance of PrunePath, LTE, and Wanda on four NLU tasks across different sparsity levels in Figure~\ref{fig:nlu_results}. Overall, PrunePath provides the strongest sparsity--accuracy trade-off on three of the four tasks and remains competitive on MNLI. Wanda is effective only under mild sparsity and degrades rapidly as the sparsity increases. LTE is more stable than Wanda, but its performance drops earlier than PrunePath at high sparsity. These results indicate that PrunePath better preserves NLU accuracy over a broad FFN sparsity range.

\subsubsection{Performance on NLG Tasks}

\begin{figure}[t]
    \centering
    \includegraphics[width=0.49\columnwidth]{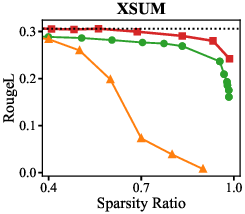} 
    \includegraphics[width=0.49\columnwidth]{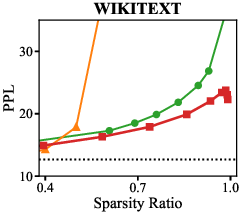} 
    \includegraphics[width=1.0\columnwidth]{Figures/legendbase.eps}
    \caption{NLG results with GPT-2 Medium. We report ROUGE-L($\uparrow$) on XSum and PPL ($\downarrow$) on WikiText.}
    \label{fig:nlg_results}
\end{figure}

Compared to NLU tasks, NLG tasks are more sensitive to sparse activation and therefore are more challenging. 
Figure~\ref{fig:nlg_results} shows the performance of the three pruning methods for the GPT-2 Medium model over XSum and WikiText, measured by ROUGE-L and perplexity (PPL), respectively.

PrunePath again achieves the best sparsity--quality trade-off. Wanda degrades very fast as sparsity increases, with ROUGE-L dropping quickly on XSum and PPL increasing rapidly on WikiText. LTE is more robust than Wanda but still shows noticeable degradation at high sparsity. In contrast, PrunePath preserves XSum generation quality over a wider sparsity range and substantially suppresses the PPL growth on WikiText, demonstrating stronger robustness on NLG tasks.

\subsubsection{Performance on Instruction Tuning Tasks}
We further evaluate whether PrunePath generalizes to instruction-tuned LLMs. We apply PrunePath to Qwen2-7B and perform supervised fine-tuning on Tulu-v2, followed by evaluation on MMLU. Compared with task-specific NLU and NLG fine-tuning, instruction tuning presents an additional routing calibration challenge. In early experiments, we observed that entropy minimization alone often produced highly sharp softmax distributions but also large positive router logits. This made many sigmoid aggregation weights close to one, even for experts that would later be removed when lowering $\tau$, increasing the mismatch between all-expert and sparse execution.

To address this issue, we add a gate-magnitude regularizer for Qwen2-7B SFT:
\begin{equation}
    \mathcal{L}_{\mathrm{gate}}
    =
    \frac{1}{BE}
    \sum_{i=1}^{B}
    \sum_{j=1}^{E}
    \mathrm{sigmoid}(g_{ij}),
\end{equation}
where $g_{ij}$ is the router logit for token $i$ and expert $j$. The final instruction-tuning objective becomes
\begin{equation}
    \mathcal{L}
    =
    \mathcal{L}_{\mathrm{task}}
    +
    \eta \mathcal{L}_{\mathrm{ent}}
    +
    \lambda \mathcal{L}_{\mathrm{bal}}
    +
    \gamma \mathcal{L}_{\mathrm{gate}}.
\end{equation}
This regularizer discourages uniformly large sigmoid gates and better aligns the softmax-based expert selection with the sigmoid-scaled expert aggregation. Table~\ref{tab:gate_reg_ablation} shows that this regularizer improves the
sparsity--accuracy frontier on the Qwen2-7B MMLU benchmark. In particular, it enables higher
FFN sparsity with less accuracy degradation, suggesting that it reduces the
selection--aggregation mismatch caused by large sigmoid gates.

\begin{table}[ht!]
\centering
\small
\caption{Effect of gate-magnitude regularization on the Qwen2-7B MMLU sparsity--accuracy frontier.}
\setlength{\tabcolsep}{4pt}
\begin{tabular}{lccc}
\toprule
Setting & \(\tau\) & Sparsity ($\%$) & MMLU ($\%$)\\
\midrule
w/o \(\mathcal{L}_{\mathrm{gate}}\) & 0.95 & 24.81 & 62.53 \\
w/  \(\mathcal{L}_{\mathrm{gate}}\) & 0.97 & 34.70 & 61.58 \\
w/o \(\mathcal{L}_{\mathrm{gate}}\) & 0.90 & 40.83 & 56.28 \\
w/  \(\mathcal{L}_{\mathrm{gate}}\) & 0.94 & 48.97 & 58.50 \\
\bottomrule
\end{tabular}
\label{tab:gate_reg_ablation}
\end{table}

\begin{figure}
    \centering
    \includegraphics[width=\linewidth]{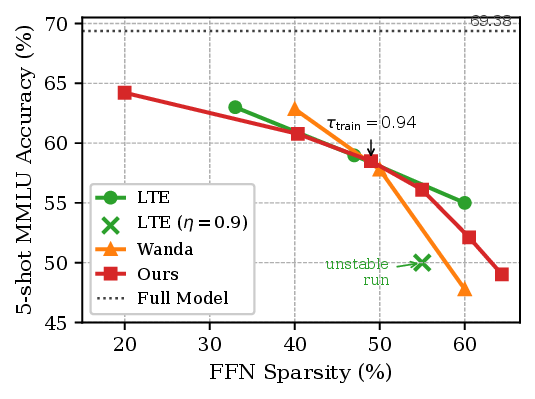}
    \caption{5-shot MMLU accuracy of Qwen2-7B after Tulu-v2 SFT. PrunePath uses a single
    checkpoint trained with $\tau_{\mathrm{train}}=0.94$, and other points are
    obtained by varying only inference-time $\tau$. 
    }
    \label{fig:instruction_mmlu}
\end{figure}

Figure~\ref{fig:instruction_mmlu} reports 5-shot MMLU accuracy under different
FFN sparsity levels. PrunePath uses a single checkpoint trained with
\(\tau_{\mathrm{train}}=0.94\), and all sparsity levels are obtained by varying
\(\tau\) only at inference time. Wanda, calibrated on C4~\cite{dodge2021documenting}, is competitive below 50\% sparsity, showing that static pruning remains effective under mild pruning budgets. However, its accuracy drops sharply under more aggressive sparsity. At higher sparsity levels, PrunePath extends a single checkpoint to 60\%+ sparsity with a smooth degradation curve, while preserving the threshold-sweep property. The main LTE frontier uses $\eta\in\{0.3,0.5,0.7\}$ for its efficiency loss, and its highest-sparsity point corresponds to $\eta=0.7$. We additionally mark a higher-$\eta$ LTE run ($\eta=0.9$), which shows an abrupt MMLU drop, indicating sensitivity to the sparsity-control coefficient. We further discuss how the mismatch between instruction-tuning data and pretraining-like calibration data may affect PrunePath in Appendix~\ref{app:instruction_calibration}.

\subsubsection{Performance on SFT Generative Models}
\begin{figure}[t]
    \centering
    \includegraphics[width=0.43\textwidth]{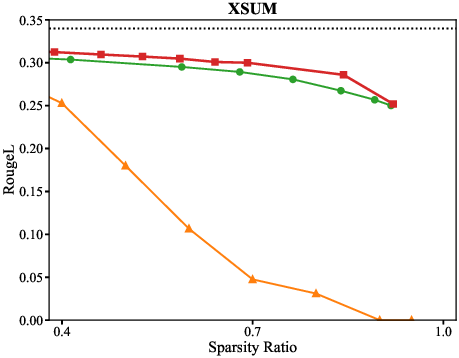}
    \includegraphics[width=\columnwidth]{Figures/legendbase.eps}
    \caption{XSum ROUGE-L of the SFT Pangu-1B model under varying FFN activation sparsity.}
    \label{fig:sft_pangu_results}
\end{figure}

We extend our evaluation to a supervised fine-tuned (SFT) model. Due to the tighter parameter dependencies introduced by supervised task adaptation, SFT models typically present a more challenging scenario for efficiency-aware methods. We benchmark Pangu-1B model on the XSum dataset, and Figure \ref{fig:sft_pangu_results} illustrates the performance trends across varying FFN sparsity levels.

All evaluated methods experience a gradual performance decline as sparsity increases, underscoring the higher sensitivity of task-adapted generation models to sparse execution. Even in this stricter scenario, PrunePath consistently maintains a clear advantage over both Wanda and LTE baselines across the entire evaluated sparsity range. Specifically, Wanda degrades steadily from the outset, failing to preserve acceptable generation quality at higher sparsity. LTE is more robust than Wanda but still tracks consistently below our curve. In contrast, PrunePath effectively dampens the rate of performance loss throughout all levels, demonstrating stronger generalization and robustness on task-adapted generative models.

\subsection{Ablation Study}

\subsubsection{Effect of Expert Initialization Strategy}

We study how the choice of weights used for expert clustering affects PrunePath. Specifically, we compare two initialization strategies: 
(1) clustering FFN neurons using the pretrained weights, and
(2) clustering FFN neurons using the downstream fine-tuned weights.
Fine-tuned weights are expected to provide more task-adapted neuron representations, while pre-trained weights offer a more generic initialization that does not rely on downstream adaptation before expert construction.

\begin{figure}[t]
    \centering
    \includegraphics[width=0.49\columnwidth]{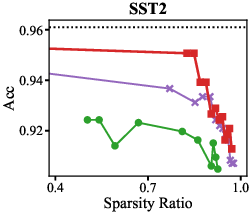}
    \includegraphics[width=0.49\columnwidth]{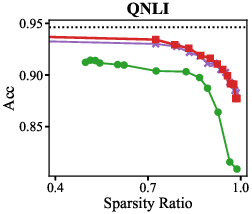}
    \includegraphics[width=1.0\columnwidth]{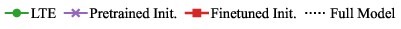}
    \caption{Effect of expert initialization on SST-2 and QNLI with RoBERTa-large.}
    \label{fig:ablation_init}
\end{figure}

Figure~\ref{fig:ablation_init} shows the results on SST-2 and QNLI with RoBERTa-large. Clustering with fine-tuned weights can yield the best sparsity--accuracy trade-off, confirming that task-adapted weights provide better priors for constructing experts. Importantly, even when experts are initialized from pre-trained weights, PrunePath remains competitive and still outperforms LTE over different sparsity levels. This indicates that PrunePath's advantage is not solely due to favorable expert initialization, but also comes from its cumulative-mass routing and progressive sparsity-path training.

\subsubsection{Top-$k$ Reconstruction Analysis}

We further analyze whether PrunePath learns a more compact expert ranking than LTE. Since the two methods use different training procedures, we first define the reference and sparse checkpoints used in this analysis. For LTE, the Stage-1 checkpoint corresponds to its soft-router mode, where all experts are evaluated and weighted by sigmoid router scores; the Stage-2 checkpoint corresponds to the hard-routing mode. For PrunePath, the Stage-1 checkpoint corresponds to the all-expert warm-up setting with $\tau=1.05$, while the Stage-2 checkpoint corresponds to the sparse routing regime with $\tau<1$. We select LTE and PrunePath checkpoints such that their Stage-1 task performance is comparable and their Stage-2 sparsity levels after adaptation are matched.

Given 100 WikiText prompts of length 512, we feed the inputs into the Stage-2 checkpoints and force both methods to retain only the top-$k$ experts ranked by their own routers at each FFN layer, with $k \in \{1,2,4,\ldots,128\}$. For each method, we compute the mean squared error (MSE) between the resulting top-$k$ FFN output and its corresponding Stage-1 all-expert output. We report the MSE over all FFN layers as a measure of how well the learned expert ranking reconstructs the high-fidelity Stage-1 representation under a fixed expert budget.

Figure~\ref{fig:mse_comparison} shows that PrunePath achieves substantially lower reconstruction error than LTE under small expert budgets. For example, at Top-1 and Top-8, PrunePath reduces the average MSE to approximately 3.5 and 3.3, respectively, compared with 4.9 and 4.1 for LTE. As $k$ increases, both methods approach their Stage-1 references and the gap narrows. These results suggest that cumulative-mass routing can place more informative experts earlier in the activation set than independent sigmoid thresholding.

\subsection{Single-Checkpoint Dynamic Sparsity}

A key practical advantage of PrunePath is that a single checkpoint can support multiple inference-time sparsity targets. To verify this, we take one GPT-2 Medium checkpoint on WikiText from our sparsity path, corresponding to the operating point $\tau=0.80$, and vary only the inference-time threshold $\tau$ \textbf{without further fine-tuning}. This produces a range of activation sparsity levels from the same checkpoint. We compare this training-free threshold sweep with LTE, which requires training separate checkpoints for different sparsity targets.

\begin{figure}[t]
    \centering
    \includegraphics[width=0.49\textwidth]{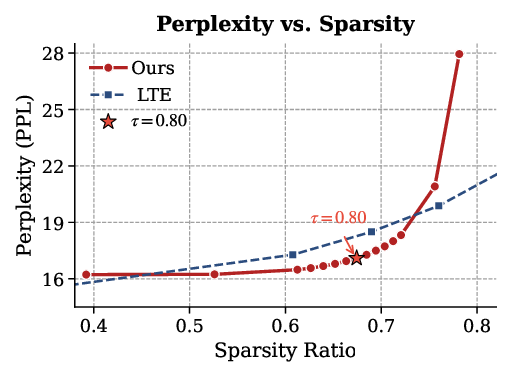}
    \caption{Inference-time $\tau$ sweep using one GPT-2 Medium checkpoint on WikiText.}
    \label{fig:ablation_tau_vs_lte}
\end{figure}

Figure~\ref{fig:ablation_tau_vs_lte} shows the resulting sparsity--perplexity trade-off. Adjusting $\tau$ smoothly controls activation sparsity, and in a broad practical sparsity range, the single PrunePath checkpoint achieves lower PPL than individually trained LTE models. This demonstrates that the learned sparsity path provides a reusable inference-time efficiency knob. However, PPL increases rapidly at extremely high sparsity, indicating that further fine-tuning is necessary when targeting very aggressive sparsity levels.

\begin{figure}
    \centering
    \includegraphics[width=\linewidth]{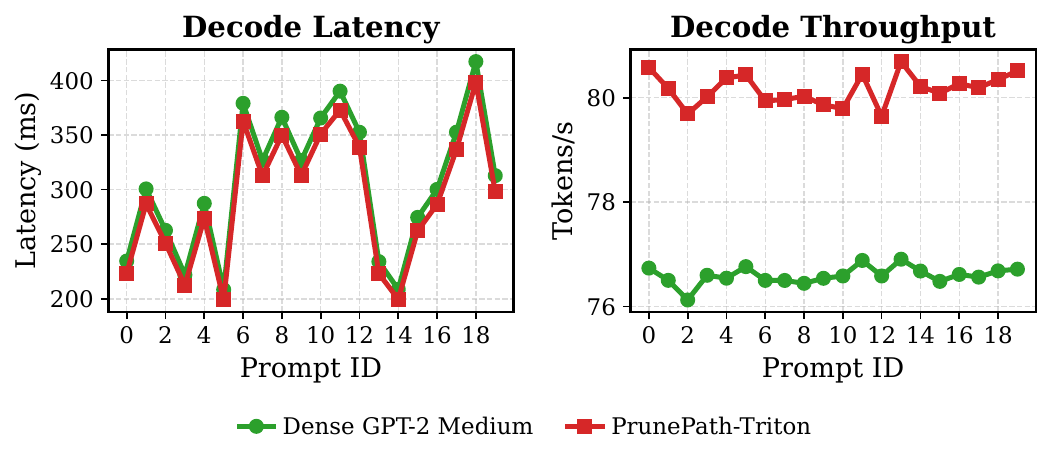}
    \caption{Per-prompt decode-only latency and throughput with KV-cache.}
    \label{fig:decode_latency_throughput}
\end{figure}

\begin{table*}[h!]
\centering
\small
\caption{
Overall decode-only inference efficiency with KV-cache on 20 XSum validation prompts.
}
\begin{tabular}{lcccc}
\toprule
\textbf{Method} 
& \textbf{Peak VRAM (MB) $\downarrow$} 
& \textbf{Decode Latency (ms) $\downarrow$} 
& \textbf{Step Latency (ms) $\downarrow$}
& \textbf{Decode TP (tok/s) $\uparrow$} \\
\midrule
Dense GPT-2 Medium 
& 2116.30 $\pm$ 12.63
& 306.15 $\pm$ 62.30
& 13.05
& 76.60 \\
PrunePath-Triton 
& 1736.88 $\pm$ 12.02
& 292.59 $\pm$ 59.78
& 12.48
& 80.15 \\
\midrule
Improvement 
& 17.93\%
& 4.45\%
& 4.45\%
& 1.046$\times$ \\
\bottomrule
\end{tabular}
\label{tab:decode_efficiency}
\end{table*}
\subsection{Triton-Accelerated Sparse FFN Inference}
\label{sec:triton_inference}

To translate PrunePath's structured FFN sparsity into practical inference benefits, we implement Triton sparse FFN kernels for GPT-2 Medium. After MoEfication, neurons assigned to the same expert are stored contiguously, allowing each selected expert to be evaluated as a dense block rather than by scattered neuron indexing. The inference path consists of lightweight routing, selected-expert evaluation, and output reduction.
Our prototype focuses on KV-cache autoregressive decoding. Although prefill is
functionally supported, it requires multi-token routing, sorting, dispatch, and
expert-wise accumulation, and is less optimized. We therefore report decode-only
latency as the main speed metric, measuring only the single-token decoding loop
after prompt prefill constructs the KV cache.

Specifically, we first run prompt prefill outside the timing region to construct the KV-cache, and then measure only the subsequent single-token decoding loop. Following HuggingFace generation, the first generated token is produced from the prefill logits; therefore, for a target summary length of $T$, we time $T-1$ decode steps. 

Table~\ref{tab:decode_efficiency} shows that PrunePath-Triton reduces peak GPU
memory by 17.9\% and improves decode-only latency by 4.4\%, increasing decoding
throughput from 76.60 to 80.15 tokens/s. The latency gain is moderate because
attention, KV-cache access, LM head computation, and generation-control
overheads are unaffected by FFN sparsity. Figure~\ref{fig:decode_latency_throughput}
further shows consistent per-prompt latency and throughput improvements.

\section{Conclusion}
We presented PrunePath, a budget-adaptive structured sparsification framework
for FFN layers in language models. PrunePath builds on MoEfication and replaces
independent expert-wise thresholding with cumulative-mass expert activation,
introducing a token-level probability budget and a direct inference-time
sparsity knob from a single checkpoint. Across NLU, NLG, and instruction-tuning evaluations, PrunePath achieves a favorable
sparsity--performance trade-off compared with static pruning and prior
MoEfication-based routing methods. Additional analyses show that PrunePath
learns more compact expert rankings and supports flexible single-checkpoint
threshold sweeping. Finally, our Triton implementation demonstrates that the
resulting structured sparsity can yield practical memory savings and measurable
decode-time speedups. These results suggest that cumulative-mass expert
activation is a simple and effective path toward deployment-friendly sparse
FFN inference.

\section*{Limitations}
\label{sec:limitations}

PrunePath provides a flexible mechanism for token-adaptive structured FFN sparsity, but it also has several limitations. 
First, cumulative-mass expert activation introduces routing overhead, since sorting expert probabilities and computing cumulative mass are less hardware-friendly than fixed top-$k$ or simple thresholding. Therefore, the realized latency gain can be smaller than the theoretical FFN computation reduction.
Second, our Triton implementation is mainly optimized for KV-cache decoding. Although we support prefill for functional completeness, the current prefill path requires multi-token routing, dispatch, sorting, and expert-wise accumulation, and remains less optimized. We thus report decode-only latency as the main speed metric and leave prefill optimization to future work.
Finally, while we evaluate RoBERTa-large, GPT-2 Medium, Pangu-1B, and Qwen2-7B, validation on larger 10B+ models and production-scale serving settings remains future work.

\newpage


\bibliography{custom}

@article{
    zheng2024learn,
    title={Learn to be efficient: Build structured sparsity in large language models},
    author={Zheng, Haizhong and Bai, Xiaoyan and Liu, Xueshen and Mao, Zhuoqing Morley and Chen, Beidi and Lai, Fan and Prakash, Atul},
    journal={Advances in Neural Information Processing Systems},
    volume={37},
    pages={101969--101991},
    year={2024}
}

@article{liu2019roberta,
  title={Roberta: A robustly optimized bert pretraining approach},
  author={Liu, Yinhan and Ott, Myle and Goyal, Naman and Du, Jingfei and Joshi, Mandar and Chen, Danqi and Levy, Omer and Lewis, Mike and Zettlemoyer, Luke and Stoyanov, Veselin},
  journal={arXiv preprint arXiv:1907.11692},
  year={2019}
}

@inproceedings{han2016deep,
  author       = {Song Han and
                  Huizi Mao and
                  William J. Dally},
  editor       = {Yoshua Bengio and
                  Yann LeCun},
  title        = {Deep Compression: Compressing Deep Neural Network with Pruning, Trained
                  Quantization and Huffman Coding},
  booktitle    = {International Conference on Learning Representations},
  year         = {2016},
}

@inproceedings{zhang2022moefication,
  title={Moefication: Transformer feed-forward layers are mixtures of experts},
  author={Zhang, Zhengyan and Lin, Yankai and Liu, Zhiyuan and Li, Peng and Sun, Maosong and Zhou, Jie},
  booktitle={Findings of the Association for Computational Linguistics: ACL 2022},
  pages={877--890},
  year={2022}
}

@article{vaswani2017attention,
  title={Attention is all you need},
  author={Vaswani, Ashish and Shazeer, Noam and Parmar, Niki and Uszkoreit, Jakob and Jones, Llion and Gomez, Aidan N and Kaiser, {\L}ukasz and Polosukhin, Illia},
  journal={Advances in Neural Information processing systems},
  volume={30},
  year={2017}
}

@article{chowdhery2023palm,
  title={Palm: Scaling language modeling with pathways},
  author={Chowdhery, Aakanksha and Narang, Sharan and Devlin, Jacob and Bosma, Maarten and Mishra, Gaurav and Roberts, Adam and Barham, Paul and Chung, Hyung Won and Sutton, Charles and Gehrmann, Sebastian and others},
  journal={Journal of machine learning research},
  volume={24},
  number={240},
  pages={1--113},
  year={2023}
}

@article{kaplan2020scaling,
  title={Scaling laws for neural language models},
  author={Kaplan, Jared and McCandlish, Sam and Henighan, Tom and Brown, Tom B and Chess, Benjamin and Child, Rewon and Gray, Scott and Radford, Alec and Wu, Jeffrey and Amodei, Dario},
  journal={arXiv preprint arXiv:2001.08361},
  year={2020}
}

@article{bradley2000constrained,
  title={Constrained k-means clustering},
  author={Bradley, Paul S and Bennett, Kristin P and Demiriz, Ayhan},
  journal={Microsoft Research, Redmond},
  volume={20},
  number={0},
  pages={0},
  year={2000}
}

@article{brown2020language,
  title={Language models are few-shot learners},
  author={Brown, Tom and Mann, Benjamin and Ryder, Nick and Subbiah, Melanie and Kaplan, Jared D and Dhariwal, Prafulla and Neelakantan, Arvind and Shyam, Pranav and Sastry, Girish and Askell, Amanda and others},
  journal={Advances in Neural Information Processing Systems},
  volume={33},
  pages={1877--1901},
  year={2020}
}

@misc{yang2024qwen2technicalreport,
      title={Qwen2 Technical Report}, 
      author={An Yang and Baosong Yang and Binyuan Hui and Bo Zheng and Bowen Yu and Chang Zhou and Chengpeng Li and Chengyuan Li and Dayiheng Liu and Fei Huang and Guanting Dong and Haoran Wei and Huan Lin and Jialong Tang and Jialin Wang and Jian Yang and Jianhong Tu and Jianwei Zhang and Jianxin Ma and Jianxin Yang and Jin Xu and Jingren Zhou and Jinze Bai and Jinzheng He and Junyang Lin and Kai Dang and Keming Lu and Keqin Chen and Kexin Yang and Mei Li and Mingfeng Xue and Na Ni and Pei Zhang and Peng Wang and Ru Peng and Rui Men and Ruize Gao and Runji Lin and Shijie Wang and Shuai Bai and Sinan Tan and Tianhang Zhu and Tianhao Li and Tianyu Liu and Wenbin Ge and Xiaodong Deng and Xiaohuan Zhou and Xingzhang Ren and Xinyu Zhang and Xipin Wei and Xuancheng Ren and Xuejing Liu and Yang Fan and Yang Yao and Yichang Zhang and Yu Wan and Yunfei Chu and Yuqiong Liu and Zeyu Cui and Zhenru Zhang and Zhifang Guo and Zhihao Fan},
      year={2024},
      eprint={2407.10671},
      archivePrefix={arXiv},
      primaryClass={cs.CL},
}

@article{li2025tpi,
  title={TPI-LLM: Serving 70B-scale LLMs efficiently on low-resource mobile devices},
  author={Li, Zonghang and Feng, Wenjiao and Guizani, Mohsen and Yu, Hongfang},
  journal={IEEE Transactions on Services Computing},
  year={2025},
  publisher={IEEE}
}

@article{dao2022flashattention,
  title={Flashattention: Fast and memory-efficient exact attention with io-awareness},
  author={Dao, Tri and Fu, Dan and Ermon, Stefano and Rudra, Atri and R{\'e}, Christopher},
  journal={Advances in Neural Information Processing Systems},
  volume={35},
  pages={16344--16359},
  year={2022}
}

@inproceedings{dao2024flashattention,
  title={Flashattention-2: Faster attention with better parallelism and work partitioning},
  author={Dao, Tri},
  booktitle={International Conference on Learning Representations},
  volume={2024},
  pages={35549--35562},
  year={2024}
}

@article{zhang2023h2o,
  title={H2o: Heavy-hitter oracle for efficient generative inference of large language models},
  author={Zhang, Zhenyu and Sheng, Ying and Zhou, Tianyi and Chen, Tianlong and Zheng, Lianmin and Cai, Ruisi and Song, Zhao and Tian, Yuandong and R{\'e}, Christopher and Barrett, Clark and others},
  journal={Advances in Neural Information Processing Systems},
  volume={36},
  pages={34661--34710},
  year={2023}
}

@inproceedings{wang2022finding,
  title={Finding skill neurons in pre-trained transformer-based language models},
  author={Wang, Xiaozhi and Wen, Kaiyue and Zhang, Zhengyan and Hou, Lei and Liu, Zhiyuan and Li, Juanzi},
  booktitle={Proceedings of the 2022 Conference on Empirical Methods in Natural Language Processing},
  pages={11132--11152},
  year={2022}
}

@article{xu2025xattention,
  title={Xattention: Block sparse attention with antidiagonal scoring},
  author={Xu, Ruyi and Xiao, Guangxuan and Huang, Haofeng and Guo, Junxian and Han, Song},
  journal={arXiv preprint arXiv:2503.16428},
  year={2025}
}

@article{beltagy2020longformer,
  title={Longformer: The long-document transformer},
  author={Beltagy, Iz and Peters, Matthew E and Cohan, Arman},
  journal={arXiv preprint arXiv:2004.05150},
  year={2020}
}

@article{zheng2024sglang,
  title={Sglang: Efficient execution of structured language model programs},
  author={Zheng, Lianmin and Yin, Liangsheng and Xie, Zhiqiang and Sun, Chuyue and Huang, Jeff and Yu, Cody H and Cao, Shiyi and Kozyrakis, Christos and Stoica, Ion and Gonzalez, Joseph E and others},
  journal={Advances in Neural Information Processing Systems},
  volume={37},
  pages={62557--62583},
  year={2024}
}

@inproceedings{kwon2023efficient,
  title={Efficient memory management for large language model serving with pagedattention},
  author={Kwon, Woosuk and Li, Zhuohan and Zhuang, Siyuan and Sheng, Ying and Zheng, Lianmin and Yu, Cody Hao and Gonzalez, Joseph and Zhang, Hao and Stoica, Ion},
  booktitle={Proceedings of the 29th symposium on operating systems principles},
  pages={611--626},
  year={2023}
}

@inproceedings{katharopoulos2020transformers,
  title={Transformers are rnns: Fast autoregressive transformers with linear attention},
  author={Katharopoulos, Angelos and Vyas, Apoorv and Pappas, Nikolaos and Fleuret, Fran{\c{c}}ois},
  booktitle={International Conference on Machine Learning},
  pages={5156--5165},
  year={2020},
  organization={PMLR}
}

@inproceedings{frantar2023sparsegpt,
  title={Sparsegpt: Massive language models can be accurately pruned in one-shot},
  author={Frantar, Elias and Alistarh, Dan},
  booktitle={International conference on machine learning},
  pages={10323--10337},
  year={2023},
  organization={PMLR}
}

@article{lecun1989optimal,
  title={Optimal brain damage},
  author={LeCun, Yann and Denker, John and Solla, Sara},
  journal={Advances in Neural Information Processing Systems},
  volume={2},
  year={1989}
}

@inproceedings{sun2024simple,
  title={A simple and effective pruning approach for large language models},
  author={Sun, Mingjie and Liu, Zhuang and Bair, Anna and Kolter, Zico},
  booktitle={International Conference on Learning Representations},
  volume={2024},
  pages={4942--4964},
  year={2024}
}

@article{mishra2021accelerating,
  title={Accelerating sparse deep neural networks},
  author={Mishra, Asit and Latorre, Jorge Albericio and Pool, Jeff and Stosic, Darko and Stosic, Dusan and Venkatesh, Ganesh and Yu, Chong and Micikevicius, Paulius},
  journal={arXiv preprint arXiv:2104.08378},
  year={2021}
}

@inproceedings{sst2,
    title = {Recursive Deep Models for Semantic Compositionality Over a Sentiment Treebank},
    author = {Socher, Richard  and
      Perelygin, Alex  and
      Wu, Jean  and
      Chuang, Jason  and
      Manning, Christopher D.  and
      Ng, Andrew  and
      Potts, Christopher},
    booktitle = {Proceedings of the 2013 Conference on Empirical Methods in Natural Language Processing},
    year = {2013},
    pages = {1631--1642},
}

@InProceedings{mnli,
  author = "Williams, Adina
            and Nangia, Nikita
            and Bowman, Samuel",
  title = "A Broad-Coverage Challenge Corpus for
           Sentence Understanding through Inference",
  booktitle = "Proceedings of the 2018 Conference of
               the North American Chapter of the
               Association for Computational Linguistics:
               Human Language Technologies, Volume 1 (Long
               Papers)",
  year = "2018",
  publisher = "Association for Computational Linguistics",
  pages = "1112--1122",
}

@inproceedings{qnli,
  author = {Rajpurkar, Pranav and Zhang, Jian and Lopyrev, Konstantin and Liang, Percy},
  title = {{SQ}u{AD}: 100,000+ Questions for Machine Comprehension of Text},
  booktitle = {Proceedings of EMNLP},
  year = {2016},
  publisher = {Association for Computational Linguistics},
  pages = {2383--2392},
}

@inproceedings{mrpc,
  title={Automatically constructing a corpus of sentential paraphrases},
  author={Dolan, William B and Brockett, Chris},
  booktitle={Proceedings of the International Workshop on Paraphrasing},
  year={2005}
}

@article{gpt2,
  title={Language Models are Unsupervised Multitask Learners},
  author={Radford, Alec and Wu, Jeff and Child, Rewon and Luan, David and Amodei, Dario and Sutskever, Ilya},
  year={2019}
}

@misc{pangu,
      title={Pangu Embedded: An Efficient Dual-system LLM Reasoner with Metacognition}, 
      author={Hanting Chen and Yasheng Wang and Kai Han and Dong Li and Lin Li and Zhenni Bi and Jinpeng Li and Haoyu Wang and Fei Mi and Mingjian Zhu and Bin Wang and Kaikai Song and Yifei Fu and Xu He and Yu Luo and Chong Zhu and Quan He and Xueyu Wu and Wei He and Hailin Hu and Yehui Tang and Dacheng Tao and Xinghao Chen and Yunhe Wang},
      year={2025},
      eprint={2505.22375},
      archivePrefix={arXiv},
      primaryClass={cs.CL}
}

@inproceedings{xsum,
  title={Don’t give me the details, just the summary! topic-aware convolutional neural networks for extreme summarization},
  author={Narayan, Shashi and Cohen, Shay B and Lapata, Mirella},
  booktitle={Proceedings of the 2018 Conference on Empirical Methods in Natural Language Processing},
  pages={1797--1807},
  year={2018}
}

@misc{wikitext,
      title={Pointer Sentinel Mixture Models},
      author={Stephen Merity and Caiming Xiong and James Bradbury and Richard Socher},
      year={2016},
      eprint={1609.07843},
      archivePrefix={arXiv},
      primaryClass={cs.CL}
}

@misc{tulu,
      title={Camels in a Changing Climate: Enhancing LM Adaptation with Tulu 2}, 
      author={Hamish Ivison and Yizhong Wang and Valentina Pyatkin and Nathan Lambert and Matthew Peters and Pradeep Dasigi and Joel Jang and David Wadden and Noah A. Smith and Iz Beltagy and Hannaneh Hajishirzi},
      year={2023},
      eprint={2311.10702},
      archivePrefix={arXiv},
      primaryClass={cs.CL}
}

@article{mmlu1,
      title={Measuring Massive Multitask Language Understanding},
      author={Dan Hendrycks and Collin Burns and Steven Basart and Andy Zou and Mantas Mazeika and Dawn Song and Jacob Steinhardt},
      journal={Proceedings of the International Conference on Learning Representations (ICLR)},
      year={2021}
}

@article{mmlu2,
    title={Aligning AI With Shared Human Values},
    author={Dan Hendrycks and Collin Burns and Steven Basart and Andrew Critch and Jerry Li and Dawn Song and Jacob Steinhardt},
    journal={Proceedings of the International Conference on Learning Representations (ICLR)},
    year={2021}
}

@article{liu2024deepseek,
  title={Deepseek-v3 technical report},
  author={Liu, Aixin and Feng, Bei and Xue, Bing and Wang, Bingxuan and Wu, Bochao and Lu, Chengda and Zhao, Chenggang and Deng, Chengqi and Zhang, Chenyu and Ruan, Chong and others},
  journal={arXiv preprint arXiv:2412.19437},
  year={2024}
}

@article{jiang2024mixtral,
  title={Mixtral of experts},
  author={Jiang, Albert Q and Sablayrolles, Alexandre and Roux, Antoine and Mensch, Arthur and Savary, Blanche and Bamford, Chris and Chaplot, Devendra Singh and Casas, Diego de las and Hanna, Emma Bou and Bressand, Florian and others},
  journal={arXiv preprint arXiv:2401.04088},
  year={2024}
}

@article{shazeer2020glu,
  title={Glu variants improve transformer},
  author={Shazeer, Noam},
  journal={arXiv preprint arXiv:2002.05202},
  year={2020}
}

@inproceedings{
Holtzman2020The,
title={The Curious Case of Neural Text Degeneration},
author={Ari Holtzman and Jan Buys and Li Du and Maxwell Forbes and Yejin Choi},
booktitle={International Conference on Learning Representations},
year={2020},
url={https://openreview.net/forum?id=rygGQyrFvH}
}

@inproceedings{tillet2019triton,
  title={Triton: an intermediate language and compiler for tiled neural network computations},
  author={Tillet, Philippe and Kung, Hsiang-Tsung and Cox, David},
  booktitle={Proceedings of the 3rd ACM SIGPLAN International Workshop on Machine Learning and Programming Languages},
  pages={10--19},
  year={2019}
}

@inproceedings{dodge2021documenting,
  title={Documenting large webtext corpora: A case study on the colossal clean crawled corpus},
  author={Dodge, Jesse and Sap, Maarten and Marasovi{\'c}, Ana and Agnew, William and Ilharco, Gabriel and Groeneveld, Dirk and Mitchell, Margaret and Gardner, Matt},
  booktitle={Proceedings of the 2021 conference on empirical methods in natural language processing},
  pages={1286--1305},
  year={2021}
}

\newpage
\appendix

\section{Effect of Calibration Data Distribution}
\label{app:instruction_calibration}
This result should be interpreted in light of the calibration distribution.
Tulu-v2 provides instruction-style supervision, whereas MMLU probes broad
knowledge acquired during pretraining. This mismatch is particularly relevant
for PrunePath because its sparse execution path is determined by a learned
softmax routing distribution: the entropy objective encourages sharp expert
rankings on the SFT data, but experts that are rare or underrepresented in
Tulu-v2 may still be important for MMLU domains. Consequently, routers calibrated
only on Tulu-v2 may not fully capture the expert-usage patterns needed across
the broader pretraining distribution. These observations suggest that a more
pretraining-like LM calibration stage, e.g., on C4-style corpora, would provide a more direct signal for knowledge-preserving sparse execution.


\end{document}